\documentclass{article}
\usepackage{spconf,amsmath,graphicx}
\usepackage{enumitem}
\usepackage{booktabs}
\usepackage{adjustbox}
\usepackage{amsfonts}
\usepackage{float}
\usepackage{url}
\usepackage{svg}
\setlist{nosep, leftmargin=14pt}
\usepackage{etoolbox}
\AtBeginEnvironment{thebibliography}{%
  \small
  \setlength{\itemsep}{0pt}%
  \setlength{\parsep}{0pt}%
  \setlength{\parskip}{0pt}%
  \setlength{\baselineskip}{9pt}%
}

\title{The APL Doesn't Fall Far From The Tree: \\ Unsupervised Artefact Prototype Learning in Histopathology}

\name{Emilia Agasi$^{1}$, Charlie Gourley$^{2}$, C. Simon Herrington$^{2}$, Robert L. Hollis$^{2}$ and T. Ian Simpson$^{1}$}
\address{$^{1}$ School of Informatics, University of Edinburgh \\ $^{2}$ CRUK Scotland Centre, Institute of Genetics and Cancer, University of Edinburgh}

\begin{document}
\maketitle

\begin{abstract}
Histopathology foundation models provide rich representations of tissue morphology, but technical artefacts in whole slide images introduce noise that may impact diagnostic accuracy. We introduce Artefact Prototype Learning (APL), a novel approach to detecting these artefacts that exploits the geometry of pre-trained embedding spaces without manual labels or fine-tuning. Rather than treating artefacts as isolated outliers, APL computes prototypes of out-of-distribution patches and uses them to distinguish artefact-associated patterns from tissue. Evaluation on a single-centre dataset shows that APL outperforms prior methods GrandQC and HistoART in detecting artefacts such as blur, dark spots, slide edges and pen marks, but struggles with tissue folds and blood. These results show that frozen pre-trained embeddings already carry information that allows artefact detection without expensive labels or vast computational resources.
\end{abstract}


\section{Introduction}
\label{sec:intro}

Digitised whole slide images (WSIs) capture the morphology and spatial architecture of many cancers, providing key diagnostic information. Foundation models (FMs) extract transferable representations from these gigapixel images by self-supervised learning (SSL) on large corpora of WSIs \cite{lu2024avisionlanguage, chen2024uni, xu2024gigapath}. These models have proven useful in many downstream tasks such as ovarian cancer subtyping \cite{breen2025comprehensive}. An often neglected issue, however, are technical artefacts present in WSIs such as pen marks, dust particles and tissue folds. When included during feature extraction, these regions are encoded alongside valid tissue and may reduce diagnostic accuracy \cite{weng2024grandqc, kahaki2026histoart}.

Existing quality control (QC) approaches typically operate on the raw gigapixel image. Traditional methods based on intensity thresholds, edge detection, colour profiles or textures can be sensitive to variations in tissue appearance and scanning protocols, and struggle to generalise across different artefact types. Supervised deep learning approaches address some of these limitations: GrandQC \cite{weng2024grandqc}, for example, uses a pixel-wise multi-class classifier that has been incorporated into existing pipelines \cite{zhang2025standardizing}. By contrast, HistoART \cite{kahaki2026histoart} performs artefact detection directly at patch level by fine-tuning the UNI model \cite{chen2024uni}, reaching high accuracy across scanners and tissue types. These approaches however, rely on labelled data. More recently, DiffusionQC \cite{wang2026diffusionqc} framed artefact detection as an outlier detection problem by learning the distribution of artefact-free images without pixel-level annotations. While this reduces the burden of per-pixel labelling, what constitutes an \emph{“outlier”} may differ per cohort due to differences in acquisition protocols.

\begin{figure}
    \centering
    \includegraphics[width=1\linewidth]{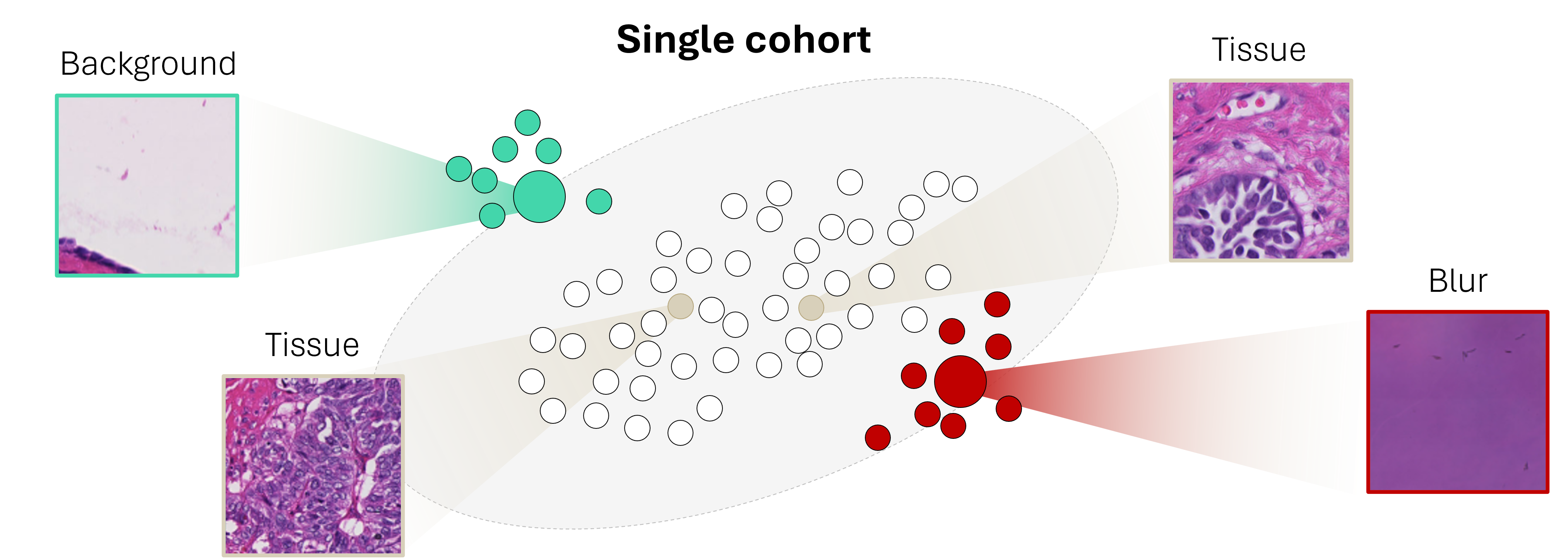}
    \caption{\textbf{Intuition behind Artefact Prototype Learning:} Artefacts naturally cluster in embedding space.}
    \label{fig:first}
    \vspace{-.5cm}
\end{figure}

Pre-trained histopathology encoders offer an opportunity to exploit this variability by mapping image patches onto a high-dimensional latent space, where similarities between embeddings reflect similarities in visual features. We ask whether these representations encode features that distinguish artefacts from tissue within a cohort, despite not being explicitly trained for artefact detection (Figure \ref{fig:first}). Building on this idea, we introduce \emph{\textbf{Artefact Prototype Learning (APL)}}, an unsupervised approach that leverages embedding space topology to identify artefacts in WSIs. We use unsupervised clustering of embeddings to identify visually related groups of patches and define a prototype as a representative embedding of each cluster, which we then use to characterise similar patches. We make the following contributions:
\begin{enumerate}
    \item APL, an unsupervised Artefact Prototype Learning method for WSI QC. Code is available at \url{https://zenodo.org/records/23889154}~\cite{emilia_agasi_2026_23889154}.
    \item We demonstrate that geometric properties of patches can be exploited to identify artefacts in embedding space.
    \item We evaluate APL on two ovarian cancer cohorts.
\end{enumerate}

\begin{figure*}[!t]
    \centering
    \includegraphics[width=1\linewidth]{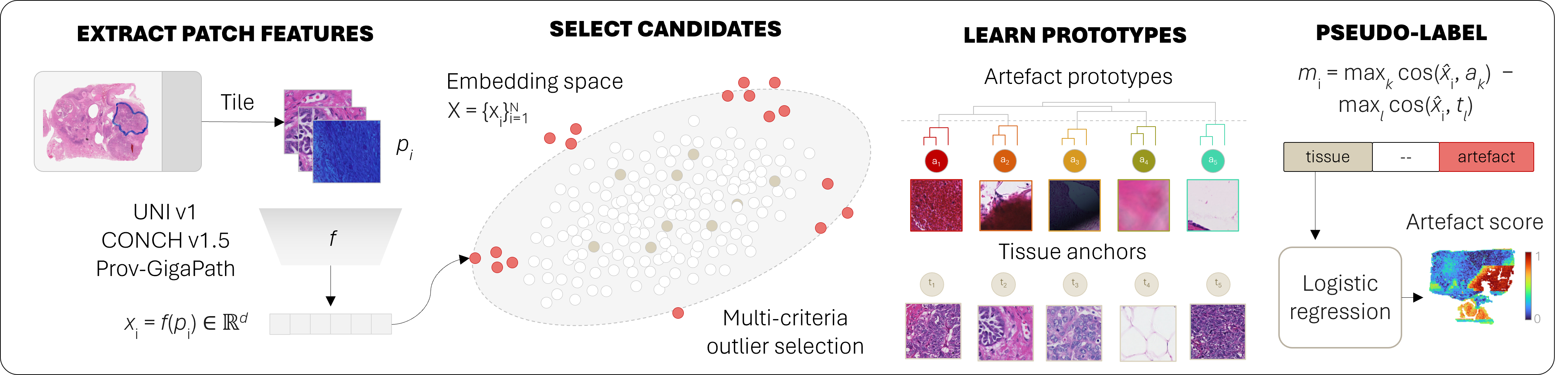}
    \vspace{-5mm}
    \caption{\textbf{Artefact prototype learning.} Overview of the APL method.}
    \label{fig:method}
    \vspace{-.20cm}
\end{figure*}

\section{Method}
\label{sec:format}

\subsection{Dataset}
We obtain haematoxylin and eosin (H\&E) stained, formalin-fixed paraffin-embedded WSIs for two separate cohorts. The first is a proprietary dataset from the Edinburgh Cancer Centre (N=414) scanned on Hamamatsu at $\sim$0.23~$\mu$m/pixel and 40$\times$ magnification \cite{hollis2022multiomic}. The second is TCGA-OV, a multi-institutional cohort (N=107) of FFPE samples from 11 tissue source sites predominantly in the United States, scanned on Aperio at 40$\times$ ($\sim$0.25~$\mu$m/pixel) \cite{cancer2011integrated}.

We pre-process raw WSIs using the TRIDENT tool \cite{zhang2025standardizing}, first segmenting tissue regions using HEST \cite{jaume2024hest}, then tiling the resulting tissue masks into $256\times256$ and $512\times512$ pixel (px) patches at $20\times$ magnification according to the native input resolution of the encoders. After ImageNet normalisation, each patch $p_i$ is mapped by encoder $f$ to an embedding $\mathbf{x}_i^{(f)}=f(p_i)\in\mathbb{R}^{d_f}$. We use CONCH v1.5 \cite{lu2024avisionlanguage}, UNI v1 \cite{chen2024uni} and Prov-GigaPath \cite{xu2024gigapath}, which produce embeddings of dimensions 768, 1024 and 1536, respectively.

\subsection{Unsupervised prototype learning}
For each cohort and encoder, we pool patch embeddings across WSIs into a common embedding space (Fig.~\ref{fig:method}). Let $\mathbf{x}_i$ denote the embedding of patch $p_i$, and let $\hat{\mathbf{x}}_i=\mathbf{x}_i/\|\mathbf{x}_i\|_2$ denote the L2-normalised embedding. We identify candidate artefacts based on three complimentary assumptions about their location in latent space:
\begin{enumerate}
    \item \emph{Angular distance}: Artefacts deviate from the mean embedding direction $\boldsymbol{\mu} = \frac{1}{N}\sum_{i=1}^{N}\hat{\mathbf{x}}_i$. We quantify this deviation as $\theta_i = \arccos\big(\cos(\hat{\mathbf{x}}_i, \boldsymbol{\mu})\big)$.
    \item \emph{Local density}: They are located in sparse regions in latent space. We measure this as the mean angular distance from each patch to its $k=20$ nearest neighbours.
    \item \emph{Mahalanobis distance:} They are unlikely under the distribution and therefore have a large Mahalanobis distance, $d_M(\mathbf{x}_i) = \sqrt{(\mathbf{x}_{i} - \boldsymbol{\mu})^T \mathbf{\Sigma}^{-1} (\mathbf{x}_{i} - \boldsymbol{\mu})}$, computed on the raw embeddings in a 50-component PCA space. Because L2-normalisation changes the covariance we use the raw embeddings.
\end{enumerate}

For candidate patches ranked in the top 1\% by at least two criteria, we apply average-linkage hierarchical clustering using cosine distance and take the L2-normalised mean of each cluster’s embeddings as its artefact prototype $\mathbf{a}_k$. We also define a number of tissue prototypes $\mathbf{t}_l$ by k-means clustering ($k=8$) of patches with $d_M(\mathbf{x}_i) \le \text{med}(d_M)$. The intuition for defining several tissue prototypes is the morphological heterogeneity in WSIs.

\subsection{Prototype-guided pseudo-labelling}
For each patch we calculate a margin by subtracting its maximum similarity to a tissue prototype from its maximum similarity to an artefact prototype given by $m_i = \max_k \cos(\hat{\mathbf{x}}_i, \mathbf{a}_k) - \max_l \cos(\hat{\mathbf{x}}_i, \mathbf{t}_l)$. Patches with $m_i > 0.1$ are labelled as artefact and patches with $m_i < 0.02$ as tissue. Patches with margins between 0.02 and 0.10 are left unlabelled. We train an L2-regularised, class-balanced binary logistic regression on the pseudo-labelled patches, which yields patch-wise artefact probabilities.

\subsection{Setup}
We evaluate our method against GrandQC (MPP25, $7\times$ model) \cite{weng2024grandqc} and HistoART (UNI-based) \cite{kahaki2026histoart} on held-out WSIs, annotated in QuPath \cite{bankhead2017qupath}. Patches are labelled as artefact if $>25\%$\,px in a patch are annotated as blur, dark spot, blood, pen mark, fold or slide edge, or if $>75\%$\,px are background. The final randomly sampled holdout set for our primary cohort comprises 15 WSIs or 76,571 512\,px and 294,661 256\,px patches with an artefact prevalence of 11.7\% and 11.4\% respectively. For TCGA-OV, the holdout set comprises 10 WSI with 127,370 256px patches (prevalence 9.1\%). Holdout slides are excluded from fitting (candidate selection and prototype learning) steps. APL operates on precomputed embeddings and runs on CPU.
To compare UNI and Prov-GigaPath embedding space geometry, we compute linear centred kernel alignment (CKA) and mKNN ($k=10$) on the same held-out 256\,px patches using null calibration ($K=200$, $\alpha=0.05$) as in Gröger et al. \cite{groger2026revisiting}. We permute patches within slides and report the median across five subsamples of 2,000 patches.

\section{Results}
\label{sec:results}

\textbf{APL detects artefacts without manual training labels.} On 15 held-out WSIs from a single-centre cohort, the UNI-based APL classifier reached an AUROC of 0.90 and AUPRC of 0.75 (\textit{Table \ref{tab:results}}). The fine-tuned HistoART model achieved a similar AUROC of 0.91 and an AUPRC of 0.69. GrandQC yielded lower AUROC and AUPRC across two patch sizes (0.66–0.67 and 0.25–0.27, respectively). AUPRC is the more informative given low artefact prevalence. Prototype-similarity margin alone, without the pseudo-label-trained classifier, reached a similar AUPRC of 0.73.
\vspace{-3mm}
\begin{table}[ht!]
    \centering
    \caption{Artefact detection performance in 15 held-out slides (artefact prevalence 10.6\% at 512\,px, 11.4\% at 256\,px), with 95\% slide-bootstrap CI. GrandQC predicts pixel masks and is scored on both patch grids. Best in bold. $^\dagger$Fine-tuned UNI.}
    \label{tab:results}
    \footnotesize
    \setlength{\tabcolsep}{4pt}           
    \renewcommand{\arraystretch}{1.2}
    \newcommand{\ci}[1]{{\scriptsize(#1)}}
    \begin{tabular}{@{}lllcc@{}}
    \toprule
    Method & Encoder & Patch & AUROC & AUPRC \\
    \midrule
    \multicolumn{5}{@{}l}{\emph{Existing approaches (supervised)}} \\
    GrandQC~\cite{weng2024grandqc} & -- & 512 & 0.67 \ci{0.59--0.77} & 0.27 \ci{0.13--0.52} \\
    GrandQC~\cite{weng2024grandqc} & -- & 256 & 0.66 \ci{0.59--0.74} & 0.25 \ci{0.13--0.46} \\
    HistoART~\cite{kahaki2026histoart} & UNI$^\dagger$ & 256 & \textbf{0.91} \ci{0.86--0.95} & 0.69 \ci{0.54--0.83} \\
    \midrule
    \multicolumn{5}{@{}l}{\emph{Artefact Prototype Learning (APL) (unsupervised)}} \\
    Margin & UNI & 256 & 0.90 \ci{0.84--0.94} & 0.73 \ci{0.56--0.86} \\
    Classifier & UNI & 256 & 0.90 \ci{0.85--0.94} & \textbf{0.75} \ci{0.60--0.86} \\
    \addlinespace
    Margin & Prov-GigaPath & 256 & 0.84 \ci{0.76--0.90} & 0.59 \ci{0.38--0.77} \\
    Classifier & Prov-GigaPath & 256 & 0.86 \ci{0.79--0.92} & 0.67 \ci{0.48--0.82} \\
    \addlinespace
    Margin & CONCH & 512 & 0.89 \ci{0.84--0.94} & 0.67 \ci{0.47--0.83} \\
    Classifier & CONCH & 512 & 0.84 \ci{0.77--0.90} & 0.62 \ci{0.43--0.78} \\
    \bottomrule
    \end{tabular}
    \vspace{-3pt}
\end{table}

\begin{figure}
    \centering
    \makebox[\columnwidth][c]{\includegraphics[width=1.06\columnwidth]{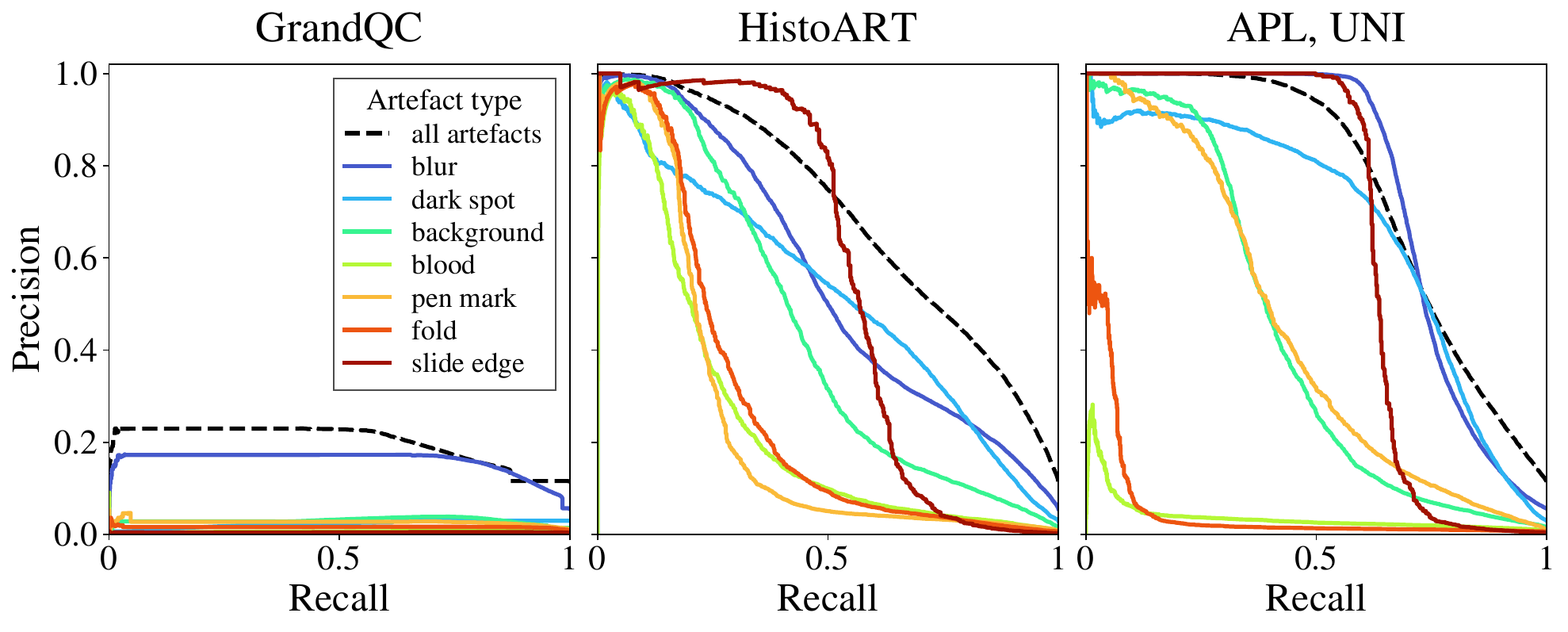}}
    \vspace{-8mm}
    \caption{Precision–recall curves per artefact for the UNI-based APL classifier, HistoART and GrandQC on held-out slides.}
    \label{fig:roc_types}
    \vspace{-15pt}
\end{figure}

\begin{figure*}
    \centering    
    \includegraphics[width=1\linewidth]{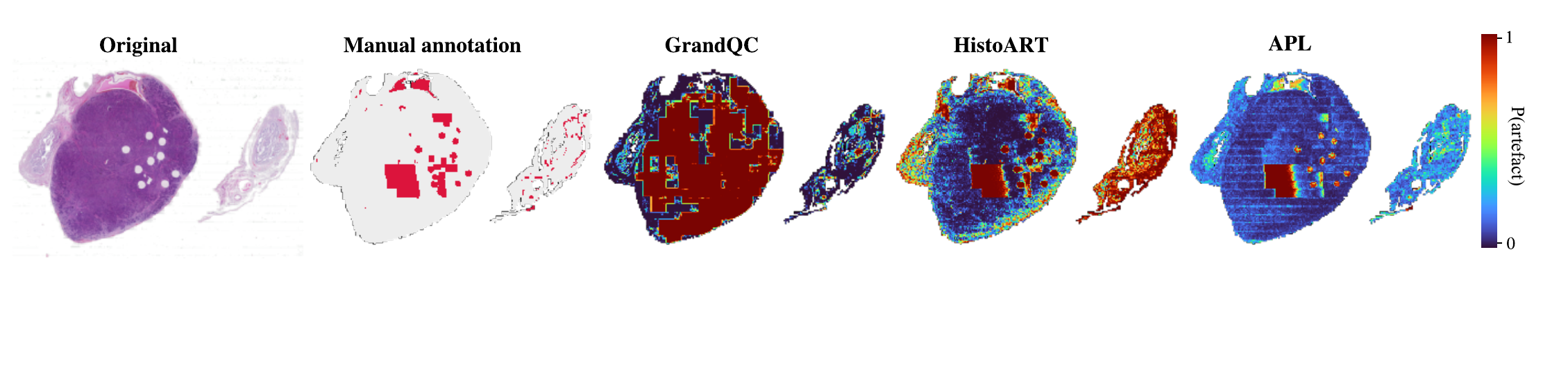}
    \vspace{-20mm}
    \caption{\textbf{Illustrative artefact probability maps for one held-out WSI.} Manual annotations mark a large blurred region and tissue-microarray punch holes. GrandQC assigns high scores across the blurred region, HistoART flags adipose tissue. APL assigns high scores to annotated artefacts.}
    \label{fig:heatmaps}
    \vspace{-5pt}
\end{figure*}

\textbf{Artefact detection performance varies by artefact type:} The UNI-based APL classifier achieved the highest AUPRC for blur (0.76), dark spots (0.67), slide edges (0.64) and pen mark detection (0.44) \textit{(Table \ref{tab:auprc_by_type})}. Folds and blood were challenging for all methods: the highest AUPRC across methods was 0.29 for folds and 0.25 for blood (Figure \ref{fig:roc_types}). In the holdout, they only constituted 0.6\% and 0.7\% of artefact patches respectively. In contrast, blur (45.8\%), dark spots (23,6\%) and background (11.3\%) were most common. Probability maps illustrate qualitative errors: GrandQC frequently overestimated out-of-focus regions, while HistoART flagged adipose tissue and APL flagged necrotic tissue (Figure \ref{fig:heatmaps}).

\begin{table}[t]
    \centering
    \vspace{-8mm}
    \caption{AUPRC by artefact type on held-out slides (Positives: target artefact,  negatives: clean tissue, chance: positive prevalence). Best per column in bold.}
    \label{tab:auprc_by_type}
    \scriptsize
    \setlength{\tabcolsep}{2.5pt}
    \begin{adjustbox}{max width=\columnwidth}
    \begin{tabular}{@{}lcccccccc@{}}
    \toprule
    Method & All & Blur & Dark spot & Background & Blood & Pen mark & Fold & Slide edge \\
    \midrule
    GrandQC~\cite{weng2024grandqc} & 0.20 & 0.16 & 0.02 & 0.03 & 0.01 & 0.03 & 0.02 & 0.00 \\
    HistoART~\cite{kahaki2026histoart} & 0.70 & 0.56 & 0.52 & \textbf{0.45} & \textbf{0.25} & 0.25 & \textbf{0.29} & 0.57 \\
    APL, UNI & \textbf{0.76} & \textbf{0.76} & \textbf{0.67} & 0.42 & 0.04 & \textbf{0.44} & 0.05 & \textbf{0.64} \\
    \midrule
    \textit{Chance} & \textit{0.115} & \textit{0.056} & \textit{0.030} & \textit{0.014} & \textit{0.008} & \textit{0.008} & \textit{0.006} & \textit{0.002} \\
    \bottomrule
    \end{tabular}
    \end{adjustbox}
    \vspace{-3mm}
\end{table}

\textbf{APL performance remains broadly consistent across encoders.} Despite differences in embedding dimension, training regimen and patch size, APL showed similar performances across encoders. UNI consistently showed the strongest overall results. Prov-GigaPath achieved lower AUROC (0.83--0.88) and AUPRC (0.58--0.65), while CONCH reached AUROC of 0.84--0.89 and AUPRC values of 0.57--0.61.
We also observe embedding space alignment in held-out patches between UNI and Prov-GigaPath, both trained using DINOv2 (256\,px patches), with a calibrated linear CKA of 0.78 capturing global alignment and mKNN of 0.52 (calibrated 0.51; within-slide null median 0.03; $p \leq 0.005$) capturing similarity in local neighbourhood structure. Alignment was similar for artefact and clean patches (calibrated CKA 0.78 vs 0.78; mKNN 0.46 vs 0.51).
Lastly, we tested our initial assumptions in held-out WSIs. Angular distance consistently distinguished artefacts from clean tissue (AUROC 0.83--0.84), predominantly identifying blurry patches, whereas Mahalanobis distance (AUROC 0.47--0.86) and kNN density (AUROC 0.54--0.81) were less consistent. 

\textbf{Artefact detection is more challenging in a multi-centre cohort:} To evaluate under more heterogeneous staining and preparation conditions, we tested APL on a multi-institutional cohort. Fitted on TCGA-OV, the UNI-based APL classifier reached an AUROC of 0.84 and an AUPRC of 0.54, below HistoART (AUROC 0.92, AUPRC 0.63), but above GrandQC (AUROC 0.73, 0.37) (Table \ref{tab:tcga_ov}). Transferring APL fitted on the single-centre cohort to TCGA-OV, reduced performance to 0.75 AUROC and 0.40 AUPRC. The underlying assumptions were also less consistent: angular distance retained some discriminative ability (AUROC 0.75), whereas Mahalanobis distance and local density performed poorly (0.39 and 0.54, respectively). Composition of artefact types also differed in this cohort: background (73\%) and blood (17\%) accounted for most artefacts. APL performed particularly well for air bubbles (AUPRC 0.86), blur (0.53) and blood (0.59).

\begin{table}[H]
    \centering
    \vspace{-5mm}
    \caption{Artefact detection on TCGA-OV. \emph{Cohort-specific}: fitted on TCGA-OV; \emph{Transferred}: fitted on single-centre.}
    \label{tab:tcga_ov}
    \footnotesize
    \setlength{\tabcolsep}{4pt}
    \renewcommand{\arraystretch}{1.2}
    \newcommand{\ci}[1]{{\scriptsize(#1)}}
    \begin{tabular}{@{}lllcc@{}}
    \toprule
    Method & Encoder & Patch & AUROC & AUPRC \\
    \midrule
    \multicolumn{5}{@{}l}{\emph{Existing approaches (supervised)}} \\
    GrandQC~\cite{weng2024grandqc} & -- & 256 & 0.73 \ci{0.57--0.83} & 0.37 \ci{0.12--0.57} \\
    HistoART~\cite{kahaki2026histoart} & UNI$^\dagger$ & 256 & \textbf{0.92} \ci{0.86--0.94} & \textbf{0.63} \ci{0.53--0.77} \\
    \midrule
    \multicolumn{5}{@{}l}{\emph{Artefact Prototype Learning (APL) (unsupervised)}} \\
    APL, cohort-specific & UNI & 256 & 0.84 \ci{0.78--0.89} & 0.54 \ci{0.31--0.70} \\
    APL, transferred     & UNI & 256 & 0.75 \ci{0.67--0.82} & 0.38 \ci{0.31--0.57} \\
    \bottomrule
    \end{tabular}
    \vspace{-1pt}
\end{table}

\section{Discussion \& Conclusion}
\label{sec:conclusion}

We present APL, a method for label-free artefact removal in embedding space. We show that we can remove artefacts with high AUPRC in a single-centre cohort and find that certain artefacts are more easily detected than others. Different encoders appear to converge to similar representations of artefacts. Lastly, we find that our approach is less consistent in multi-centre cohorts, which we assume may be due to acquisition protocol differences.
Because APL relies on embedding space topology rather than explicit definitions of artefacts, atypical tissue patterns may also be flagged, which we observe in the case of necrosis and blood, the latter is an artefact under \cite{weng2024grandqc}. These may deviate from the dominant tissue distribution but are not necessarily technical artefacts. This leads to a broader discussion on how we define artefacts and the impact of removal of certain types of patches on performance in downstream tasks.
Ultimately, APL provides a lightweight, cohort-centric approach to artefact detection. Flagged patches can then be excluded from slide-level representations to prevent artefacts from influencing downstream analyses.

\section{Compliance with ethical standards}
\label{sec:ethics}
We acquired ethical approval from South East Scotland Human Annotated Bioresource (Lothian NRS Bioresource Ethics Committee, reference 15/ES/0094-SR705 and SR752). For this retrospective study, consent was not required by the committee. This study was carried out in accordance with the principles of the Declaration of Helsinki. Additional data was acquired from the TCGA Research Network: \url{https://www.cancer.gov/tcga}.

\section{Acknowledgments}
\label{sec:acknowledgments}
This work was supported by the United Kingdom Research and Innovation (grant EP/Y030869/1) AI Centre for Doctoral Training in Biomedical Innovation at the University of Edinburgh. For the purpose of open access, the author has applied a creative commons attribution (CC BY) licence to any author accepted manuscript version arising. The authors have no relevant financial or non-financial interests to disclose. Generative AI was used to correct spelling and grammatical errors.

\bibliographystyle{IEEEbib}
\bibliography{strings,refs}

@article{lu2024avisionlanguage,
  title={A visual-language foundation model for computational pathology},
  author={Lu, Ming Y and Chen, Bowen and Williamson, Drew FK and Chen, Richard J and Liang, Ivy and Ding, Tong and Jaume, Guillaume and Odintsov, Igor and Le, Long Phi and Gerber, Georg and others},
  journal={Nature Medicine},
  pages={863–874},
  volume={30},
  year={2024},
  publisher={Nature Publishing Group}
}

@article{chen2024uni,
  title={Towards a General-Purpose Foundation Model for Computational Pathology},
  author={Chen, Richard J and Ding, Tong and Lu, Ming Y and Williamson, Drew FK and Jaume, Guillaume and Chen, Bowen and Zhang, Andrew and Shao, Daniel and Song, Andrew H and Shaban, Muhammad and others},
  journal={Nature Medicine},
  publisher={Nature Publishing Group},
  year={2024}
}

@article{xu2024gigapath,
  title={A whole-slide foundation model for digital pathology from real-world data},
  author={Xu, Hanwen and Usuyama, Naoto and Bagga, Jaspreet and Zhang, Sheng and Rao, Rajesh and Naumann, Tristan and Wong, Cliff and Gero, Zelalem and González, Javier and Gu, Yu and Xu, Yanbo and Wei, Mu and Wang, Wenhui and Ma, Shuming and Wei, Furu and Yang, Jianwei and Li, Chunyuan and Gao, Jianfeng and Rosemon, Jaylen and Bower, Tucker and Lee, Soohee and Weerasinghe, Roshanthi and Wright, Bill J. and Robicsek, Ari and Piening, Brian and Bifulco, Carlo and Wang, Sheng and Poon, Hoifung},
  journal={Nature},
  year={2024},
  publisher={Nature Publishing Group UK London}
}

@article{weng2024grandqc,
  title={GrandQC: A comprehensive solution to quality control problem in digital pathology},
  author={Weng, Zhilong and Seper, Alexander and Pryalukhin, Alexey and Mairinger, Fabian and Wickenhauser, Claudia and Bauer, Marcus and Glamann, Lennert and Bl{\"a}ker, Hendrik and Lingscheidt, Thomas and Hulla, Wolfgang and others},
  journal={Nature Communications},
  volume={15},
  number={1},
  pages={10685},
  year={2024},
  publisher={Nature Publishing Group UK London}
}

@article{hollis2022multiomic,
  title={Multiomic characterization of high-grade serous ovarian carcinoma enables high-resolution patient stratification},
  author={Hollis, Robert L and Meynert, Alison M and Michie, Caroline O and Rye, Tzyvia and Churchman, Michael and Hallas-Potts, Amelia and Croy, Ian and McCluggage, W Glenn and Williams, Alistair RW and Bartos, Clare and others},
  journal={Clinical Cancer Research},
  volume={28},
  number={16},
  pages={3546--3556},
  year={2022},
  publisher={American Association for Cancer Research}
}

@article{cancer2011integrated,
  title={Integrated genomic analyses of ovarian carcinoma},
  author={Cancer Genome Atlas Research Network and others},
  journal={Nature},
  volume={474},
  number={7353},
  pages={609},
  year={2011}
}

@inproceedings{jaume2024hest,
    author = {Guillaume Jaume and Paul Doucet and Andrew H. Song and Ming Y. Lu and Cristina Almagro-Perez and Sophia J. Wagner and Anurag J. Vaidya and Richard J. Chen and Drew F. K. Williamson and Ahrong Kim and Faisal Mahmood},
    title = {HEST-1k: A Dataset for Spatial Transcriptomics and Histology Image Analysis},
    booktitle = {Advances in Neural Information Processing Systems},
    year = {2024},
    month = dec,
}

@article{kahaki2026histoart,
  title={HistoART: Histopathology artifact detection based on large foundation model},
  author={Kahaki, Seyed and Webber, Alexander and Zamzmi, Ghada and Subbaswamy, Adarsh and Deshpande, Rucha and Badano, Aldo},
  journal={Neural Computing and Applications},
  volume={38},
  number={18},
  pages={753},
  year={2026},
  publisher={Springer}
}

@article{bankhead2017qupath,
  title={QuPath: Open source software for digital pathology image analysis},
  author={Bankhead, Peter and Loughrey, Maurice B and Fern{\'a}ndez, Jos{\'e} A and Dombrowski, Yvonne and McArt, Darragh G and Dunne, Philip D and McQuaid, Stephen and Gray, Ronan T and Murray, Liam J and Coleman, Helen G and others},
  journal={Scientific reports},
  volume={7},
  number={1},
  pages={16878},
  year={2017},
  publisher={Nature Publishing Group}
}

@article{zhang2025standardizing,
  title={Accelerating Data Processing and Benchmarking of AI Models for Pathology},
  author={Zhang, Andrew and Jaume, Guillaume and Vaidya, Anurag and Ding, Tong and Mahmood, Faisal},
  journal={arXiv preprint arXiv:2502.06750},
  year={2025}
}

@article{groger2026revisiting,
  title={Revisiting the platonic representation hypothesis: An aristotelian view},
  author={Gr{\"o}ger, Fabian and Wen, Shuo and Brbi{\'c}, Maria},
  journal={arXiv preprint arXiv:2602.14486},
  year={2026}
}

@article{breen2025comprehensive,
  title={A comprehensive evaluation of histopathology foundation models for ovarian cancer subtype classification},
  author={Breen, Jack and Allen, Katie and Zucker, Kieran and Godson, Lucy and Orsi, Nicolas M and Ravikumar, Nishant},
  journal={NPJ Precision Oncology},
  volume={9},
  number={1},
  pages={33},
  year={2025},
  publisher={Nature Publishing Group UK, London}
}

@inproceedings{wang2026diffusionqc,
  title={DiffusionQC: Artifact Detection and Quality Control in Histopathology Images via Diffusion Model},
  author={Wang, Zhenzhen and Zhou, Zhongliang and Wen, Zhuoyu and Kook, Jeong Hwan and Wojcik, John B and Kang, John},
  booktitle={2026 IEEE 23rd International Symposium on Biomedical Imaging (ISBI)},
  pages={1--5},
  year={2026},
  organization={IEEE}
}

@misc{emilia_agasi_2026_23889154,
  author       = {Emilia Agasi},
  title        = {emilia-agasi/artefact\_prototype\_learning: APL
                   v0.1.0
                  },
  month        = nov,
  year         = 2026,
  publisher    = {Zenodo},
  version      = {0.1.0},
  doi          = {10.5281/zenodo.23889154},
  url          = {https://doi.org/10.5281/zenodo.23889154},
}

\end{document}